# Identifying Ventricular Arrhythmias and Their Predictors by Applying Machine Learning Methods to Electronic Health Records in Patients With Hypertrophic Cardiomyopathy (HCM-VAr-Risk Model)

Moumita Bhattacharya, PhD[a], Dai-Yin Lu, MD[b,c,d], Shibani M. Kudchadkar, MD[b], Gabriela Villarreal Greenland, MD[b,e], Prasanth Lingamaneni, MD[b], Celia P. Corona-Villalobos, MD[b,f], Yufan Guan[b], Joseph E. Marine[b], Jeffrey E. Olgin, MD[e], Stefan Zimmerman, MD[f], Theodore P. Abraham, MD[b,e], Hagit Shatkay, PhD[a,b,1,*], and Maria Roselle Abraham, MD[b,e,1,*]

Clinical risk stratification for sudden cardiac death (SCD) in hypertrophic cardiomyopathy (HC) employs rules derived from American College of Cardiology Foundation/American Heart Association (ACCF/AHA) guidelines or the HCM Risk-SCD model (C-index ~0.69), which utilize a few clinical variables. We assessed whether data-driven machine learning methods that consider a wider range of variables can effectively identify HC patients with ventricular arrhythmias (VAr) that lead to SCD. We scanned the electronic health records of 711 HC patients for sustained ventricular tachycardia or ventricular fibrillation. Patients with ventricular tachycardia or ventricular fibrillation (n = 61) were tagged as VAr cases and the remaining (n = 650) as non-VAr. The 2-sample $t$ test and information gain criterion were used to identify the most informative clinical variables that distinguish VAr from non-VAr; patient records were reduced to include only these variables. Data imbalance stemming from low number of VAr cases was addressed by applying a combination of over- and under-sampling strategies. We trained and tested multiple classifiers under this sampling approach, showing effective classification. We evaluated 93 clinical variables, of which 22 proved predictive of VAr. The ensemble of logistic regression and naïve Bayes classifiers, trained based on these 22 variables and corrected for data imbalance, was most effective in separating VAr from non-VAr cases (sensitivity = 0.73, specificity = 0.76, C-index = 0.83). Our method (HCM-VAr-Risk Model) identified 12 new predictors of VAr, in addition to 10 established SCD predictors. In conclusion, this is the first application of machine learning for identifying HC patients with VAr, using clinical attributes. Our model demonstrates good performance (C-index) compared with currently employed SCD prediction algorithms, while addressing imbalance inherent in clinical data.   © 2019 The Authors. Published by Elsevier Inc. This is an open access article under the CC BY-NC-ND license. (http://creativecommons.org/licenses/by-nc-nd/4.0/) (Am J Cardiol 2019;123:1681−1689)

[a]Department of Computer and Information Sciences, Computational Biomedicine Lab, University of Delaware, Newark, Delaware; [b]Hypertrophic Cardiomyopathy Center of Excellence, Johns Hopkins University, Baltimore, Maryland; [c]Division of Cardiology, Taipei Veterans General Hospital, Taipei, Taiwan; [d]Institute of Public Health, National Yang-Ming University, Taipei, Taiwan; [e]Division of Cardiology, Hypertrophic Cardiomyopathy Center of Excellence, University of California San Francisco, San Francisco, California; and [f]Department of Radiology, Johns Hopkins University, Baltimore, Maryland. Manuscript received November 13, 2018; revised manuscript received and accepted February 11, 2019.

Funding: This work was funded in part by NSF IIS EAGER grant #1650851, an award from the John Taylor Babbitt (JTB) foundation (Chatham, New Jersey) and startup funds from the UCSF Division of Cardiology (to MRA).  Dr. Lu was supported by the Taipei Veterans General Hospital-National Yang-Ming University Excellent Physician Scientists Cultivation Program, no. 104-V-A-005.

[1]These authors contributed equally to the work.
See page 1688 for disclosure information.
*Corresponding authors: Tel: (415)502-3911; fax: (415) 502-7949 (MRA); Tel: (302)831-8622 (HS).
E-mail addresses: shatkay@udel.edu (H. Shatkay), Roselle.Abraham@ucsf.edu (M.R. Abraham).

Hypertrophic cardiomyopathy (HC) is the most common genetic cardiovascular disease and a major cause of sudden cardiac death (SCD) in young individuals.[1−4] This has led to development of clinical guidelines for SCD risk stratification and implantable cardioverter defibrillator (ICD) implantation.[1,2] The guidelines are based on retrospective data and methods that rely on hand-crafted rules, or on observed associations between measurements and the medical condition to identify HC patients at high risk for SCD.[1,2,4] These methods are fragile in the face of new data and demonstrate a relatively low level of performance, reflected by a C-index of ~0.69 for the HCM Risk-SCD prediction model.[4] In this study, we use machine learning to develop and evaluate a computational method (HCM-VAr-Risk Model) that addresses data imbalance, and utilizes a set of clinical variables to identify HC patients with lethal VAr, characterized as sustained ventricular tachycardia or ventricular fibrillation (VT/VF). Machine learning methods offer the advantage of flexibility to update the model as additional clinical data becomes available.

## Methods

The HC Registry is approved by the Institutional Review Boards of the Johns Hopkins Hospital and the University of California San Francisco. Patients were enrolled in the HC Registry during their first visit to the Johns Hopkins Hypertrophic Cardiomyopathy Center of Excellence, if they met the standard diagnostic criteria for HC, namely, unexplained left ventricular hypertrophy (maximal wall thickness ≥15 mm)[2] in the absence of uncontrolled hypertension, valvular heart disease, and HC phenocopies such as amyloidosis and storage disorders.

We performed a retrospective study of all HC patients from the HC Registry, who were seen at the Johns Hopkins Hypertrophic Cardiomyopathy Center of Excellence between 2005 to 2015. Patients were followed for a mean duration of 2.86 years (median = 1.92; twenty-fifth to seventy-fifth percentile = 0.94 to 4.28 years). Clinical data including symptoms, co-morbidities, medications, history of arrhythmias, and risk factors for SCD[2] were ascertained by the examining physician (MRA and TPA) during the initial clinic visit, and during each follow-up visit. Rest and stress echocardiography (ECHO) and cardiac magnetic resonance imaging were performed as part of patients' clinical evaluation. Review of the electronic health records for clinical data was performed jointly by SMK, PL, and DYL. Analysis of ECHO and cardiac magnetic resonance imaging was performed by DYL and GVL/CPCV, respectively. All analyses were blinded to VAr outcome. (See Supplementary Section A1-2 for detailed imaging methods.)

Arrhythmic events were recorded by reviewing electrocardiogram (ECG), Holter monitor, and ICD interrogation data. VAr were defined as sustained VT (ventricular rate ≥130 beats/min/≥30 seconds duration) or VF, resulting in defibrillator shocks or antitachycardia pacing, and were confirmed by an electrophysiologist (JEM, MRA). Nonsustained VT (NSVT) was defined as ≥3 consecutive ventricular beats at a rate of ≥100 bpm/<30 seconds in duration. Patients who did not have implantable defibrillators (ICDs) were followed annually by Holter monitoring; patients with ICDs had device interrogation performed every 6 months, or more frequently if they were symptomatic or had ICD discharges.

HC patients with at least 1 episode of sustained VT or VF were labeled as VAr cases; the remainder were considered non-VAr. The computational 5-step framework (HCM-VAr-Risk Model) used for identifying patients in the VAr group is presented in Figure 1. The 5 steps are as follows: (1) preprocessing to remove variables directly correlated with VAr, and to address missing data; (2) feature selection to identify the most informative clinical variables for separating VAr from non-VAr cases; (3) association analysis to identify the degree of association between each predictor variable and the VAr class; (4) supervised machine learning for creating classifiers and performing classification; and (5) a thorough quantitative and qualitative evaluation to assess the classifier's performance.

We preprocessed the data to remove variables known to be noninformative with respect to VAr (such as visit date, patient ID) as well as descriptors of VAr (e.g. number of VT episodes, ICD shocks) and adverse outcomes (such as atrial fibrillation, heart failure, and stroke). After this step, our feature set consisted of 93 clinical variables (Supplementary Table S1). As some records were missing values, data imputation for these values was employed using a nearest neighbor approach. (See Supplementary Section B.1.1.)

Classifiers trained based on high-dimensional data often exhibit low sensitivity and specificity, as many of the features are not sufficiently informative for separating the different classes (in this case patient records indicating VAr from non-VAr records). Moreover, some features may show discriminating power within a limited dataset but do not generalize beyond the training. To reduce data dimensionality, we conducted feature selection, identifying attributes that are most informative for VAr. Notably, using fewer attributes to represent the dataset yields lower variance around learned model parameters and around performance measures across training sets and test sets, and helps avoid overfitting. Our dataset is characterized by both nominal and continuous attributes (features). Although the classification method developed throughout this work is multivariate, the feature selection step is performed in a univariate fashion. To select highly predictive nominal features, we utilized the well-known information gain criterion,[5] measuring the information gained about the VAr-class given the value assumed by the feature. For continuous features, we employed the 2 sample $t$ test under unequal variance,[6,7] testing whether the distribution of attribute values associated with VAr cases are significantly different from those associated with non-VAr. We included in the reduced feature set only those continuous attributes for which the $t$ test indicated a highly statistically significant distributional difference ($p \leq 0.05$), and the nominal attributes for which the information gain value was ≥0.002. The latter threshold was determined empirically by iteratively removing the least informative feature, 1 at a time, and conducting classification without it. The procedure was repeated until deterioration in performance was observed. The feature selection process identified 22 clinical variables as informative for VAr in HC patients (Table 1).

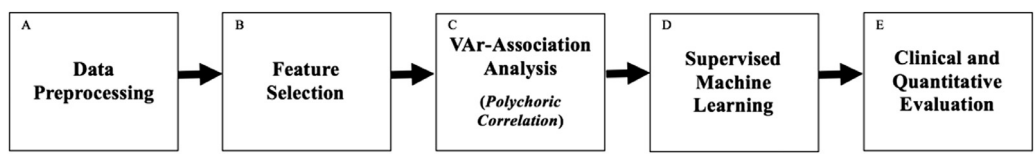

Figure 1. HCM-VAr-Risk model schematic: an overview of the framework employed to identify ventricular arrhythmia (VAr) from clinical attributes in electronic health records (HCM-VAr-Risk Model). The first step, data preprocessing, involves removing variables known to be noninformative with respect to VAr and variables associated with adverse outcomes such as heart failure, atrial fibrillation, and stroke. The second step is feature selection to identify the most informative clinical variables for separating VAr cases from non-VAr cases. As a third step, association analysis was performed to identify the degree of association between each predictor variable and the VAr class. Next, supervised machine learning was employed to create the classifiers and perform the classification. Lastly, a thorough quantitative and qualitative evaluation was conducted to assess the classifier's performance.

Many of the attributes gathered per patient are nominal rather than continuous-numerical (Supplementary Table S1). These include the predicted outcome variable itself (VAr vs non-VAr), and other items such as HC type or history of syncope. The degree of association between such nominal-valued features cannot be assessed using the standard Pearson correlation, but are quantified through the polychoric correlation,[8,9] which ranges over $[-1, 1]$, where $-1$ indicates negative association and 1 positive association.

A classifier takes as input, a vector of values (in our case, a patient's record) and assigns a probability indicating the likelihood of the patient to belong to 1 of the 2 classes: VAr versus non-VAr. We use the 22 features that effectively distinguish VAr cases from non-VAr, for representing each patient record in our dataset. Specifically, each of the 711 patients, denoted $p^i$ ($1 \leq i \leq 711$), is represented as a 22-dimensional vector, $V^i = \langle p_1^i, \ldots, p_{22}^i \rangle$, where each dimension corresponds to the clinical value obtained for the respective attribute. The classifier takes the 22 dimensional vectors $V^i$ and calculates the probability of the $i$th patient to be a VAr case, $Pr(\text{VAr} \mid V^i)$ versus its probability to be non-VAr, $(Pr(\text{non-VAr}|V^i) = 1 - Pr(\text{VAr}|V^i))$. The higher the value $Pr(\text{VAr} \mid V^i)$, the more likely the patient is to have VAr. For instance, a probability of 0.95 for VAr assigned to patient $p$ indicates a high VAr risk, whereas a probability of 0.45 indicates a much lower risk.

As a first simple baseline, we used 4 standard machine learning classification methods, namely, logistic regression, naïve Bayes, decision tree, and random forest, in an attempt to identify whether a patient has VAr. (Python scikit-learn package was used to train these 4 baseline classifiers.[10]) Trained on our highly imbalanced dataset, all standard classifiers performed poorly, failing to detect almost any VAr records (Table 3). As such, we devised a method combining over- and under-sampling together with an ensemble classifier that combines the most effective classifiers to separate

Table 1

Clinical and imaging characteristics of HC cohort stratified by presence/absence of ventricular arrhythmia (VAr)

| Variable | Ventricular arrhythmia | | p-value |
|---|---|---|---|
| | No (n = 650) | Yes (n = 61) | |
| Age (years) | 54 ± 15 | 49 ± 16 | 0.03 |
| Male | 397 (60%) | 37 (61%) | 0.9 |
| Body mass index, kg/m$^2$ | 30 ± 7 | 28 ± 5 | 0.09 |
| HC type* | | | <0.001 |
|   Non-obstructive | 186 (28%) | 34 (56%) | |
|   Labile-obstructive | 247 (38%) | 11 (18%) | |
|   Obstructive | 224 (34%) | 16 (26%) | |
| NYHA class | | | 0.7 |
|   I | 359 (55%) | 32 (53%) | |
|   II-III | 298 (45%) | 29 (47%) | |
| Angina | 258 (40%) | 21 (34%) | 0.5 |
| Family history of HC | 108 (17%) | 17 (28%) | 0.04 |
| ICD implantation | 25 (4%) | 32 (53%) | <0.001 |
| Unexplained syncope | 116 (18%) | 21 (34%) | 0.003 |
| Family history of SCD | 151 (23%) | 18 (30%) | 0.3 |
| Non-sustained VT | 0 (0%) | 11 (18%) | <0.001 |
| ECHO: Septal wall thickness ≥3 cm | 36 (6%) | 8 (13%) | 0.04 |
| ECHO: left atrial diameter (mm) | 42 ± 7 | 43 ± 8 | 0.3 |
| ECHO: maximal septal wall thickness (mm) | 20 ± 5 | 22 ± 5 | 0.001 |
| ECHO: left ventricular ejection fraction (%) | 66 ± 8 | 63 ± 9 | 0.003 |
| ECHO: E/A | 1.4 ± 0.8 | 1.5 ± 0.9 | 0.6 |
| ECHO: E/e′ | 18 ± 11 | 22 ± 16 | 0.08 |
| ECHO Peak rest LVOT gradient (mm Hg) | 31 ± 33 | 23 ± 28 | 0.04 |
| ECHO: Peak stress LVOT gradient (mm Hg) | 74 ± 54 | 44 ± 45 | <0.001 |
| CMR: LGE (% of LV mass) | 12 ± 13 | 16 ± 14 | 0.1 |
| Medications | | | |
|   Beta-blocker | 451 (69%) | 53 (87%) | 0.005 |
|   Calcium channel blocker | 184 (28%) | 14 (23%) | 0.5 |
|   RAS blockade | 161 (25%) | 12 (20%) | 0.5 |
|   Disopyramide | 24 (4%) | 3 (5%) | 0.9 |

CMR = cardiac magnetic resonance imaging; E/A = ratio of early diastolic mitral flow velocity to the late diastolic mitral flow velocity; E/e' = ratio of early diastolic mitral flow velocity to early diastolic mitral septal annulus motion velocity; ECHO = echocardiogram; HC = hypertrophic cardiomyopathy; ICD = implantable cardioverter defibrillator; LGE = late gadolinium enhancement; LVOT = left ventricular outflow tract; NYHA = New York Heart Association; RAS blockade = angiotensin-converting enzyme inhibitor, angiotensin receptor blocker; SCD = sudden cardiac death; VAr = ventricular arrhythmia; VF = ventricular fibrillation; VT = ventricular tachycardia; LV = left ventricle.

* Classification of HC was established based on LVOT/mid-cavitary gradients, as:1. Nonobstructive HC (gradients <30 mm Hg at rest and stress).2. Labile-obstructive HC (<30 mm Hg at rest, ≥30 mm Hg with stress).3. Obstructive HC (≥30 mm Hg at rest and stress).

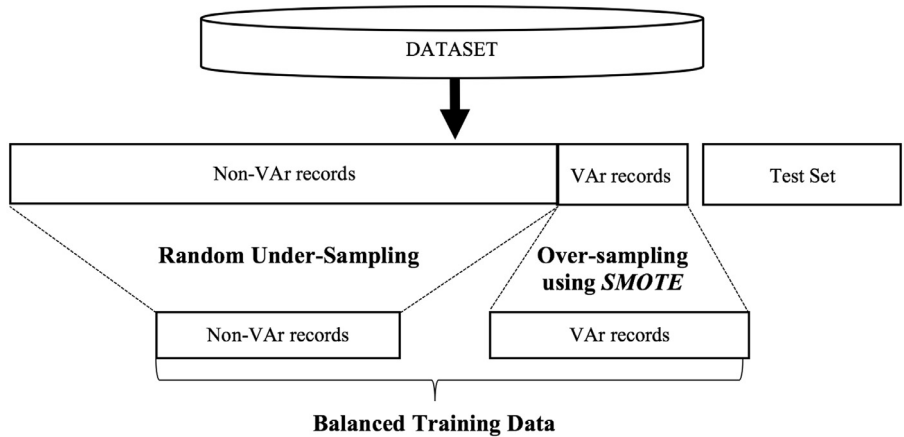

Figure 2. Methods to address data imbalance: data partitioning scheme for the combined over- and undersampling classification method. The topmost cylinder represents the entire dataset. The *rectangles* below it represent the split into test and training sets for fivefold crossvalidation. The non-VAr records (majority class) in the training set are randomly undersampled such that the non-VAr to VAr records ratio is 3:1. The VAr records (minority class) in the training set are oversampled using SMOTE to generate new minority class records, doubling the original number of minority class records. This combined under- and oversampling yields a balanced training set, containing the same number of VAr and non-VAr records.

VAr records from non-VAr records, while addressing imbalance. Figure 2 illustrates the data partitioning scheme used in the combined over- and undersampling scheme, as it was applied to the training set. A detailed description of the classification model and its testing is presented in Supplementary Section B1.2.

To address data imbalance, we also applied methods that were previously reported in the study, such as simple oversampling, simple undersampling, adaptive synthetic sampling approach[11] and meta-classification[12] and found that the performance of our combined under- and oversampling using SMOTE was superior. Hence, we report only the results obtained using our method (HCM-VAr-Risk Model) that combines under- and oversampling and compare it against the baseline classifiers.

To compare our VAr identification model (HCM-VAr-Risk Model) with contemporary clinical practice, we evaluated the performance of our model when trained on datasets represented through 2 different feature sets. These feature sets were constructed based on attributes identified as predictive of VAr in established clinical guidelines.[1,2] The first feature set comprised attributes deemed informative for SCD risk stratification in the American College of Cardiology Foundation/American Heart Association (ACCF/AHA) guidelines for HC diagnosis and treatment.[2] The second comprised attributes in the HCM Risk-SCD prediction model employed in the European Society of Cardiology (ESC) guidelines.[1,4]

To assess the performance of our HCM-VAr-Risk Model, we used several common performance measures,[5] namely, specificity, sensitivity (recall), false-negative rate (miss rate), and area under receiver-operating characteristics (ROC) curve. The first 3 measures are defined as follows:

$$Specificity = \frac{TN}{TN + FP},$$

$$Sensitivity = \frac{TP}{TP + FN},$$

$$False\ Negative\ Rate\ (Miss\ Rate) = \frac{FN}{FN + TP},$$

where TP (true positives) denotes VAr records that are correctly labeled as VAr by the classifier; TN (true negatives) denotes records that are not associated with VAr and are not assigned to this class by the classifier; FP (false positives) denotes records not associated with VAr that are misclassified by the classifier as VAr; FN (false negatives) denotes VAr records that were incorrectly labeled by the classifier as non-VAr. The ROC curve shows the true-positive rate, calculated as $\frac{TP}{TP+FN}$, as a function of false-positive rate, calculated as $\frac{FP}{FP+TN}$ (FP—false positive). Classifier performance is assessed based on the area under the ROC curve (AUC; also referred to as C-index).

## Results

Demographic and clinical features of the HC cohort are presented in Table 1; a list of all clinical variables tested in the HCM-VAr-Risk model is available in Supplementary Table S1. Our cohort consists of 711 patients with a clinical diagnosis of HC. VAr, characterized by sustained VT or VF occurred in 8% of the HC population. The VAr group was younger, had greater septal wall thickness, and was more likely to have nonobstructive hemodynamics and family history of HC. No difference was observed in the amount of late gadolinium enhancement in the left ventricle (reflecting replacement fibrosis) in the 2 groups (Table 1).

Our feature selection process identified 22 clinical variables whose values distinguish VAr from non-VAr cases within the HC population (Table 2). Of these, 11 variables were negatively correlated with VAr, whereas the remaining 11 were positive associated with VAr. We found that lower values of LVOT gradients at rest/stress, global longitudinal early diastolic strain rate, baseline systolic/diastolic BP, exercise capacity, left ventricle ejection fraction, body mass index, statin use, and age are associated with higher risk of VAr in HC. Furthermore, unexplained syncope, family history of HC or SCD, NSVT, inducible VT by noninvasive programmed stimulation, and higher values for septal hypertrophy, E/e′ and (less negative) global longitudinal

Table 2
Variables included in the HCM-VAr-Risk Model

| Variables | Type of variable | p-value | Polychoric correlation (association with VAr) |
|---|---|---|---|
| Stress LVOT gradient (mm Hg) (−) | Continuous | 0.00001 | −0.273 |
| Unexplained syncope (Presence +) | Nominal | 0.0003 | 0.264 |
| NSVT (Presence +) | Nominal | 0.0005 | 0.994 |
| **HC type** (Non-obstructive +) | Nominal | 0.001 | |
| **(1) Non-obstructive** | | | **0.366** |
| (2) Labile-obstructive | | | −0.283 |
| (3) Obstructive | | | −0.112 |
| **SBP before exercise test** (mm Hg) (−) | Continuous | 0.001 | −0.232 |
| **Global longitudinal early diastolic strain rate** (−) | Continuous | 0.001 | −0.213 |
| Maximal IVS thickness (mm) (+) | Continuous | 0.003 | 0.125 |
| **Global longitudinal systolic strain rate** (+) | Continuous | 0.003 | **0.171** |
| **Exercise time on treadmill** (−) | Continuous | 0.007 | −0.167 |
| **ECHO LVEF (%)** (−) | Continuous | 0.01 | −0.198 |
| IVS/PW ratio (+) | Continuous | 0.01 | 0.195 |
| **DBP before exercise test** (mm Hg) (−) | Continuous | 0.01 | **−0.177** |
| **METs** (−) | Continuous | 0.01 | **−0.131** |
| VT by NIPS during follow-up (Presence +) | Nominal | 0.01 | 0.667 |
| **Body mass index** (kg/m$^2$) (−) | Continuous | 0.03 | **−0.115** |
| **Global longitudinal LV systolic strain, %** (+) | Continuous | 0.03 | **0.235** |
| Age (−) | Continuous | 0.03 | −0.150 |
| LVOT gradient at rest (mm Hg) (−) | Continuous | 0.04 | −0.119 |
| Family history of SCD (Presence +) | Nominal | 0.05 | 0.097 |
| **Family history of HC** (Presence +) | Nominal | 0.06 | 0.195 |
| E/e′ (+) | Continuous | 0.06 | 0.167 |
| **Statin use** (−) | Nominal | 0.06 | **−0.052** |

DBP = diastolic blood pressure; E/e′ = ratio of early diastolic mitral flow velocity to the early diastolic mitral septal annulus velocity; HC = hypertrophic cardiomyopathy; IVS = interventricular septum; IVS/PW = ratio of maximal thickness of inter ventricular septum and maximal thickness of posterior wall of left ventricle; LV = left ventricle; LVEF = left ventricle ejection fraction; LVOT = left ventricular outflow tract; METS = metabolic equivalents; NIPS = non-invasive programmed stimulation; NSVT = nonsustained ventricular tachycardia; SBP = systolic blood pressure; SCD = sudden cardiac death.

Variables identified through feature selection (22 of the 93 variables), as highly informative of ventricular arrhythmia (VAr). Variables are listed in increasing order of their respective *p-value* (third column). Variables that have not been associated with VAr prediction in the ACC-AHA or ESC guidelines are shown in boldface.

"+" indicates that the variable has a higher value in patients with VAr, compared with non-VAr patients, and "−" indicates a lower value of the variable in patients with VAr. The rightmost column shows the value of the polychoric correlation between each variable and VAr. The *p-value* is calculated during the feature-selection process. The polychoric correlation is an additional measure used to indicate the direction of association (positive or negative correlation) between VAr and each of the variables identified as predictive of VAr by our feature selection method.

systolic strain/strain rate are positively associated with VAr. Table 2 provides a list of these predictive variables, along with the corresponding polychoric correlation and *p-values*, indicating their degree of association (or lack thereof) with VAr.

We found that combining the ensemble classifier comprising logistic regression and naïve Bayes with over- and undersampling led to higher sensitivity, higher AUC, and lower false-negative rate, compared with the 4 simple classifiers (naïve Bayes, logistic regression, decision tree, and random forest) alone (Table 3 and Figure 3). Figure 4 illustrates the C-index (AUC = 0.83) for our method (HCM-VAr-Risk Model), which assigns individualized probabilities for VAr.

Table 3
Comparison of performance between baseline classifier and classifier resulting from the HCM-VAr-Risk Model

| Performance | Baseline | HCM-VAr-Risk Model combination of over and under-sampling |
|---|---|---|
| Sensitivity | 0.20 (*0.06*) | **0.73** (*0.03*) |
| Specificity | **0.96** (*0.04*) | 0.76 (*0.04*) |
| False-negative rate (miss rate) | 0.80 (*0.06*) | **0.27** (*0.05*) |
| AUC (C-index) | 0.80 (*0.04*) | **0.83** (*0.05*) |

AUC = area under receiver-operating curve.

Comparison of performance between the simple baseline logistic regression classifier (denoted baseline) and the classifier resulting from our HCM-VAr-Risk Model, trained on datasets represented through the 22 features identified by our feature selection method. Standard deviation is shown in parentheses. The best performance attained for each measure is shown in boldface.

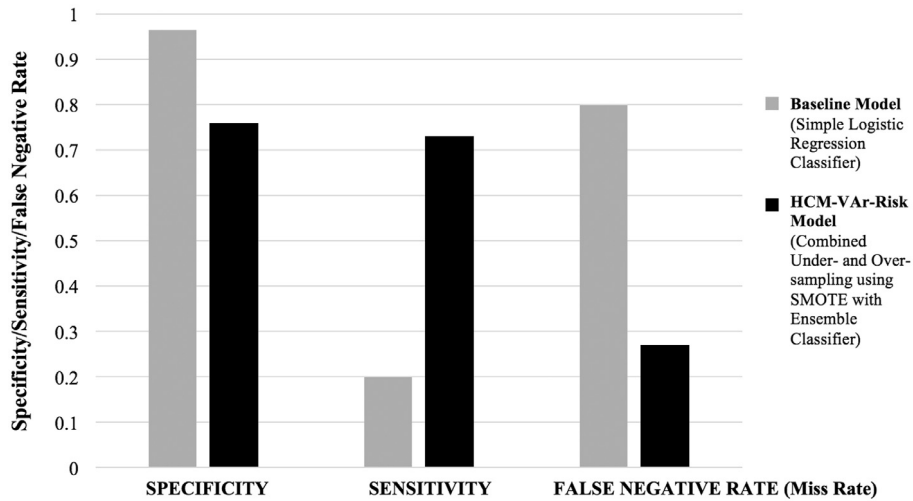

Figure 3. Comparison of performance between baseline model and HCM-VAr-Risk Model. Performance of the simple logistic regression classifier (baseline model) and the classifier resulting from our under- and oversampling scheme (HCM-VAr-Risk Model) for classifying patient records as VAr versus non-VAr. Classifiers were trained on records using the 22 most informative features to represent patients. The x-axis shows the performance measure, whereas the y-axis shows the levels of performance with respect to specificity, sensitivity, and false-negative rate.

We compared the performance of the HCM-VAr-Risk model with that reported by earlier studies, namely employing ACCF/AHA guidelines[2] and ESC guidelines (HCM Risk-SCD Model).[4] Table 4 lists 2 sets of variables (rows 1 to 2) compared with the set of 22 clinical variables identified by our HCM-VAr-Risk Model. Our model demonstrated higher specificity, sensitivity, area under ROC curve (C-index), and lower false-negative rate compared with published predictors utilized by current clinical guidelines (Table 5). Notably, the C-index attained by our model (0.83) is significantly higher than that reported in the study by O'Mahony et al[4] (0.69), which forms the basis of ESC guidelines. The dataset used in their study[4] is not publicly available and as such we could not use it for training/testing our model. Thus, we compare the performance level attained by our model with that reported in O'Mahony et al's study while applying the same evaluation metric, C-index.

We did not include adverse clinical outcomes such as atrial fibrillation, stroke, or heart failure in the list of clinical variables because our goal was to identify demographic, clinical, and imaging features that predict adverse outcomes (VAr in this case) in HC. Since HC patients with more severe cardiac phenotype would be expected to have other adverse cardiac outcomes, we examined whether atrial fibrillation and heart failure are associated with VAr in our HC cohort. We found a positive association between VAr and atrial fibrillation ($p = 0.008$ using the 2-sample $t$-test; polychoric correlation = 0.26), and between VAr and heart failure ($p = 0.039$, polychoric correlation = 0.172). Inclusion of atrial fibrillation and heart failure in our model did not increase the C-index, but led a small reduction in specificity of our model from 0.76 to 0.73.

### Discussion

This is the first application of machine learning to distinguish HC patients at high versus low risk for VAr using clinical variables, while addressing data imbalance. Although previous studies have developed data-driven models toward

Table 4

Comparison of variables and performance of the ACCF/AHA and ESC guidelines with the HCM-VAr-Risk Model, to predict SCD in HC patients

| Feature set | Variables | Performance |
| --- | --- | --- |
| *Feature set 1* (ACCF/AHA guidelines[2]) | History of NSVT, abnormal blood pressure response to exercise, family history of SCD, maximum LV wall thickness ≥3cm, unexplained syncope, family history of SCD, number of SCD risk factors | Not available |
| *Feature set 2* (ESC guidelines[1]) | Age (years), maximal LV wall thickness (mm), left atrial diameter (mm), LV outflow gradient (mm Hg), family history SCD, history of NSVT, syncope | C-index/AUC = 0.69 |
| *Feature set 3* HCM-VAr-Risk Model (our method) | Features identified by our feature-selection method (shown in Table 2) | C-index/AUC = 0.83 |

　AUC = area under receiver-operating curve; LV = left ventricle; LVOT = left ventricular outflow tract; NSVT = nonsustained ventricular tachycardia; SCD = sudden cardiac death.

　Feature sets 1 and 2 include variables identified as pertinent by 2 previous studies: feature set 1 consists of the attributes deemed as informative for identification of HC patients at high risk for SCD in the ACCF/AHA guidelines[2] for HC diagnosis and treatment. Feature set 2 consists of attributes utilized by the HCM Risk-SCD Model.[1,5] Feature set 3 denotes the 22 attributes identified as informative by our feature selection methods (HCM-VAr-Risk Model). The rightmost column shows performance, in terms of C-index/AUC, for feature sets 2 and 3. Performance level was not reported by the respective studies that identified feature set 1, and is thus not shown.

Table 5

Performance comparison of the combination of under- and oversampling logistic regression classifier, when trained on datasets represented through 2 different feature sets

|  | Feature set 1 | Feature set 2 | Feature set 3 |
| --- | --- | --- | --- |
| Sensitivity | 0.53 (*0.05*) | 0.56 (*0.03*) | **0.73** (*0.03*) |
| Specificity | 0.70 (*0.05*) | **0.76** (*0.03*) | **0.76** (*0.04*) |
| False-negative rate (miss rate) | 0.46 (*0.05*) | 0.43 (*0.04*) | **0.27** (*0.05*) |
| C-index | 0.61 (*0.07*) | 0.77 (*0.04*) | **0.83** (*0.05*) |

Feature set 1: history of nonsustained ventricular tachycardia, abnormal blood pressure response to exercise, family history of sudden cardiac death, maximum LV wall thickness ≥3 cm, and unexplained syncope.

Feature set 2: age (years), maximal LV wall thickness (mm), left atrial diameter (mm), LV outflow tract gradient (mm Hg), family history of sudden cardiac death, history of nonsustained ventricular tachycardia, and syncope.

Feature set 3: The 22 features identified as predictive of VAr by our HCM-VAr-Risk Model.

Standard deviation is shown in parentheses and the best performance attained for each measure is shown in boldface.

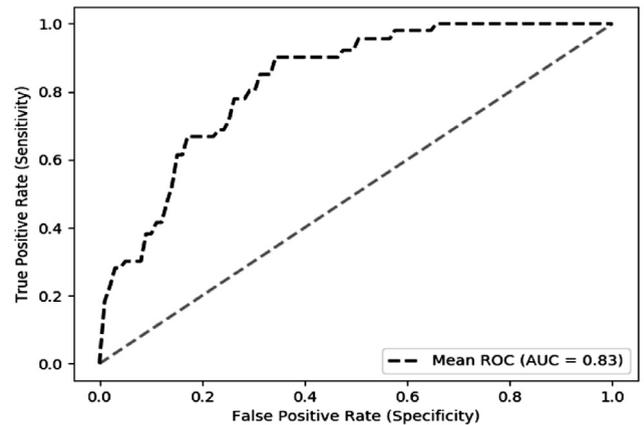

Figure 4. Receiver-operating characteristic (ROC) curve for HCM-VAr-Risk Model: ROC curve showing performance of the HCM-VAr-Risk Model using the combined under- and oversampling approach. The x-axis shows the false-positive rate, whereas the y-axis shows the true-positive rate.

identifying risk for HC, none aimed to assess risk of VAr in the HC population. Moreover, none was based just on clinical variables.[13−15] A study by Lyon et al[14] identified subgroups of HC patients displaying different electrophysiological and structural phenotypes using computational analysis of 12-lead Holter ECGs. Similarly, a study by Rahman et al[15] developed a machine learning approach to distinguish between ECG signals of HC from non-HC patients. Notably, none of the previously developed models handled class imbalance, which often occurs in the context of HC patients with high-risk versus low-risk for VAr.

In the context of risk stratification, particularly when assigning a severe event such as a VAr label to a record, high sensitivity (correct identification of cases with VAr) and low false negatives (i.e., avoiding missing actual VAr cases) are especially important. Although our method (HCM-VAr-Risk Model) outperforms the simple baseline classifiers in terms of sensitivity, false-negative rate, and C-index, it has lower specificity (correct identification of non-VAr cases)—Table 3 and Figure 4. There is usually a trade-off between sensitivity and specificity; however, a classifier that aims for high sensitivity is preferable, given the goal of accurate VAr classification. That said, having a large number of false-positives (low specificity) is clearly undesirable as it generates false alarms. Although the specificity of the simple logistic regression classifier is much higher (0.96) compared with that attained by our model (0.76), the former fails to identify almost any high-risk case (minority class), attaining a sensitivity of just 0.20. Our model's significantly higher sensitivity while retaining acceptable level of specificity demonstrates its effectiveness in correctly identifying the large majority of both VAr and non-VAr cases.

Unlike our study, which employs a data-driven feature selection method to separate HC patients with history of VAr from those without VAr, the ACCF/AHA[2] and ESC guidelines[1] incorporate a hierarchical approach that includes weighting of some variables such as family history and maximum left ventricular wall thickness in VAr risk prediction. The ESC guidelines[1] are based on a predictive 5-year risk model (HCM RISK-SCD Model)[4] that excludes patients with previous history of VF, which makes direct comparison with our cohort problematic. Another difference between our method and the ACCF/AHA guidelines[2] is the inclusion of arbiters such as complex genotypes, apical aneurysms, and late gadolinium enhancement for VAr risk prediction in the published guidelines.

Our dataset is highly imbalanced, that is, the number of non-VAr patient records greatly exceeds (by 11-fold) the number of VAr records. This is expected in HC, which is characterized by phenotypic heterogeneity and variable penetrance. Furthermore, only a minority of HC patients demonstrate adverse outcomes.[16] Learning classifiers using off-the-shelf packages on such an imbalanced dataset typically result in poor performance (Table 3), as demonstrated by the failure of logistic regression, naïve Bayes, decision tree, and random forest classifiers to correctly identify majority of the VAr cases. In contrast, our HCM-VAr-Risk Model, which directly addresses the imbalance, identified VAr cases with a sensitivity of 0.73 and a low false-negative rate of 0.27. We anticipate that our method can also be applied to other HC outcomes and to diseases that are associated with a wide spectrum of clinical phenotypes.

We identified 12 variables that have not been associated with VAr in HC by previous studies, as well as 10 established SCD risk factors (including younger age, unexplained syncope, NSVT, wall thickness, and family history of SCD)[17,18] that are utilized in the ACCF/AHA[2] and/or ESC guidelines.[1] We did not detect an association between VAr and LA diameter (ESC guidelines), abnormal BP response, or left ventricular LGE (ACCF/AHA guidelines). Presence of LGE, which reflects replacement fibrosis, is frequent in HC patients, but its utility as an independent risk factor for SCD in HC is unresolved.[19,20]

Notably, lower LVOT rest/stress gradients were associated with higher VAr risk in our study. This result is supported by a previous study that reported an association of nonobstructive HC with VAr, and of obstructive HC with atrial fibrillation, heart failure, and death.[21] The mechanism underlying the association between nonobstructive HC and VAr[22,23] could be presence of a greater degree of LV myopathy in patients with nonobstructive HC.[24] The association between statin use and VAr in HC is novel. Interestingly, statin therapy has been demonstrated to have a modest beneficial effect on SCD in the setting of CAD,[25] and to induce regression of cardiac hypertrophy and fibrosis in experimental models[26,27] of HC.

We used speckle tracking to image myocardial deformation and also performed conventional measurements of systolic and diastolic function by ECHO. Our results indicate that HC patients with VAr have a more severe cardiac phenotype, characterized by greater impairment of systolic and diastolic cardiac mechanics. It is unlikely that the impairment of cardiac mechanics is age related because younger age[28] was associated with VAr in our study. Our findings of more severe cardiac HC phenotype is supported by lower exercise capacity[29] in the VAr group, which does not appear to be related to LVOT obstruction, because patients with nonobstructive HC[24] had higher risk for VAr.

The HCM-VAr-Risk Model supports identification of patients with VAr based on their clinical record, within the realistic context of a highly imbalanced case population. This approach can be applied as part of a system alerting physicians of high or low VAr risk based on a patient record, enabling interventions such as ICD implantation in high-risk patients and avoiding ICDs in low-risk patients. Specifically, our machine learning method can assign VAr probability to any HC patient for whom values of the informative variables are available. Although we have shown very good prediction performance in the face of data imbalance, the relatively small dataset reduces the statistical power. Future work includes testing and extending the generalizability of our model using additional datasets and prospective studies.

In conclusion, this is the first application of machine learning to predict VAr in HC patients. Our machine learning-based approach demonstrates good performance (sensitivity = 0.73, specificity = 0.76, C-index = 0.83) compared with currently employed SCD prediction algorithms, while addressing the imbalance inherent in clinical data. The set of clinical attributes identified by our method includes several new predictors of VAr in HC, and indicates that patients with VAr have a more severe cardiac HC phenotype.

**Disclosures**

The authors have no conflicts of interest to disclose.

**Supplementary materials**

Supplementary material associated with this article can be found in the online version at https://doi.org/10.1016/j.amjcard.2019.02.022.

1. Authors/Task Force MembersElliott PM, Anastasakis A, Borger MA, Borggrefe M, Cecchi F, Charron P, Hagege AA, Lafont A, Limongelli G, Mahrholdt H, McKenna WJ, Mogensen J, Nihoyannopoulos P, Nistri S, Pieper PG, Pieske B, Rapezzi C, Rutten FH, Tillmanns C, Watkins H. 2014 ESC guidelines on diagnosis and management of hypertrophic cardiomyopathy: the task force for the diagnosis and management of hypertrophic cardiomyopathy of the European Society of Cardiology (ESC). *Eur Heart J* 2014;35:2733–2779.
2. American College of Cardiology Foundation/American Heart Association Task Force on PracticeAmerican Association for Thoracic SurgeryAmerican Society of EchocardiographyAmerican Society of Nuclear CardiologyHeart Failure Society of AmericaHeart Rhythm SocietySociety for Cardiovascular Angiography and InterventionsSociety of Thoracic SurgeonsGersh BJ, Maron BJ, Bonow RO, Dearani JA, Fifer MA, Link MS, Naidu SS, Nishimura RA, Ommen SR, Rakowski H, Seidman CE, Towbin JA, Udelson JE, Yancy CW. 2011 ACCF/AHA guideline for the diagnosis and treatment of hypertrophic cardiomyopathy: a report of the American College of Cardiology Foundation/American Heart Association Task Force on Practice Guidelines. *Circulation* 2011;124:e783–e831.
3. Marian AJ. On predictors of sudden cardiac death in hypertrophic cardiomyopathy. *J Am Coll Cardiol* 2003;41:994–996.
4. O'Mahony C, Jichi F, Pavlou M, Monserrat L, Anastasakis A, Rapezzi C, Biagini E, Gimeno JR, Limongelli G, McKenna WJ, Omar RZ, Elliott PM. Hypertrophic Cardiomyopathy Outcomes Investigators. A novel clinical risk prediction model for sudden cardiac death in hypertrophic cardiomyopathy (HCM risk-SCD). *Eur Heart J* 2014;35:2010–2020.
5. Murphy KP. *Machine Learning: A Probabilistic Perspective*. Cambridge: MIT Press; 2012.
6. Sokal RR, Rohlf FJ. *Introduction to Biostatistics*. 2nd ed. New York: Freeman; 1987.
7. Welch BL. The significance of the difference between two means when the population variances are unequal. *Biometrika* 1938;29:350–362.
8. Drasgow F. Polychoric and polyserial correlations. *Ency Statist Sci* 1988;7:69–74.
9. Olsson U. Maximum likelihood estimation of the polychoric correlation coefficient. *Psychometrika* 1979;44:443–460.
10. Abraham A, Pedregosa F, Eickenberg M, Gervais P, Mueller A, Kossaifi J, Gramfort A, Thirion B, Varoquaux G. Machine learning for neuroimaging with scikit-learn. *Front Neuroinform* 2014;8:14.
11. Tan AC, Gilbert D, Deville Y. Multi-class protein fold classification using a new ensemble machine learning approach. *Genome Inform* 2003;14:206–217.
12. Bhattacharya M, Jurkovitz C, Shatkay H. Assessing chronic kidney disease from office visit records using hierarchical meta-classification of an imbalanced dataset. In: 2017 IEEE International Conference on Bioinformatics and Biomedicine (BIBM); 2017. p. 663–670.
13. Baessler B, Mannil M, Maintz D, Alkadhi H, Manka R. Texture analysis and machine learning of non-contrast T1-weighted MR images in patients with hypertrophic cardiomyopathy-preliminary results. *Eur J Radiol* 2018;102:61–67.
14. Lyon A, Ariga R, Minchole A, Mahmod M, Ormondroyd E, Laguna P, de Freitas N, Neubauer S, Watkins H, Rodriguez B. Distinct ECG phenotypes identified in hypertrophic cardiomyopathy using machine learning associate with arrhythmic risk markers. *Front Physiol* 2018;9:213.
15. Rahman QA, Tereshchenko LG, Kongkatong M, Abraham T, Abraham MR, Shatkay H. Utilizing ECG-based heartbeat classification for hypertrophic cardiomyopathy identification. *IEEE Trans Nanobiosci* 2015;14:505–512.
16. Maron BJ, Rowin EJ, Casey SA, Link MS, Lesser JR, Chan RH, Garberich RF, Udelson JE, Maron MS. Hypertrophic cardiomyopathy in adulthood associated with low cardiovascular mortality with contemporary management strategies. *J Am Coll Cardiol* 2015;65:1915–1928.
17. Christiaans I, van Engelen K, van Langen IM, Birnie E, Bonsel GJ, Elliott PM, Wilde AA. Risk stratification for sudden cardiac death in hypertrophic cardiomyopathy: systematic review of clinical risk markers. *Europace* 2010;12:313–321.
18. Wang W, Lian Z, Rowin EJ, Maron BJ, Maron MS, Link MS. Prognostic implications of nonsustained ventricular tachycardia in high-risk patients with hypertrophic cardiomyopathy. *Circ Arrhythm Electrophysiol* 2017;10.
19. Prinz C, Schwarz M, Ilic I, Laser KT, Lehmann R, Prinz EM, Bitter T, Vogt J, van Buuren F, Bogunovic N, Horstkotte D, Faber L. Myocardial fibrosis severity on cardiac magnetic resonance imaging predicts sustained arrhythmic events in hypertrophic cardiomyopathy. *Can J Cardiol* 2013;29:358–363.

20. Weng Z, Yao J, Chan RH, He J, Yang X, Zhou Y, He Y. Prognostic value of LGE-CMR in HCM: a meta-analysis. *JACC Cardiovasc Imaging* 2016;9:1392–1402.
21. Lu DY, Pozios I, Haileselassie B, Ventoulis I, Liu H, Sorensen LL, Canepa M, Phillip S, Abraham MR, Abraham TP. Clinical outcomes in patients with nonobstructive, labile, and obstructive hypertrophic cardiomyopathy. *J Am Heart Assoc* 2018;7. pii: e006657. https://doi.org/10.1161/JAHA.117.006657.
22. Elliott PM. Hypertrophic cardiomyopathy: job done or work in progress? *J Am Coll Cardiol* 2016;67:1410–1411.
23. Hebl VB, Miranda WR, Ong KC, Hodge DO, Bos JM, Gentile F, Klarich KW, Nishimura RA, Ackerman MJ, Gersh BJ, Ommen SR, Geske JB. The natural history of nonobstructive hypertrophic cardiomyopathy. *Mayo Clin Proc* 2016;91:279–287.
24. Pozios I, Corona-Villalobos C, Sorensen LL, Bravo PE, Canepa M, Pisanello C, Pinheiro A, Dimaano VL, Luo H, Dardari Z, Zhou X, Kamel I, Zimmerman SL, Bluemke DA, Abraham MR, Abraham TP. Comparison of outcomes in patients with nonobstructive, labile-obstructive, and chronically obstructive hypertrophic cardiomyopathy. *Am J Cardiol* 2015;116:938–944.
25. Rahimi K, Majoni W, Merhi A, Emberson J. Effect of statins on ventricular tachyarrhythmia, cardiac arrest, and sudden cardiac death: a meta-analysis of published and unpublished evidence from randomized trials. *Eur Heart J* 2012;33:1571–1581.
26. Patel R, Nagueh SF, Tsybouleva N, Abdellatif M, Lutucuta S, Kopelen HA, Quinones MA, Zoghbi WA, Entman ML, Roberts R, Marian AJ. Simvastatin induces regression of cardiac hypertrophy and fibrosis and improves cardiac function in a transgenic rabbit model of human hypertrophic cardiomyopathy. *Circulation* 2001;104:317–324.
27. Senthil V, Chen SN, Tsybouleva N, Halder T, Nagueh SF, Willerson JT, Roberts R, Marian AJ. Prevention of cardiac hypertrophy by atorvastatin in a transgenic rabbit model of human hypertrophic cardiomyopathy. *Circ Res* 2005;97:285–292.
28. Maron BJ, Roberts WC, Epstein SE. Sudden death in hypertrophic cardiomyopathy: a profile of 78 patients. *Circulation* 1982;65:1388–1394.
29. Luo HC, Dimaano VL, Kembro JM, Hilser A, Hurtado-de-Mendoza D, Pozios I, Tomas MS, Yalcin H, Dolores-Cerna K, Mormontoy W, Aon MA, Cameron D, Bluemke DA, Stewart KJ, Russell SD, Cordova JG, Abraham TP, Abraham MR. Exercise heart rates in patients with hypertrophic cardiomyopathy. *Am J Cardiol* 2015;115:1144–1150.